\documentclass{article}

\usepackage[final]{nips_2016}

\usepackage{natbib}

\usepackage{algorithm}
\usepackage{algorithmic}

\usepackage{hyperref}
\usepackage[usenames,dvipsnames]{color}
\definecolor{dark-gray}{gray}{0.30}
\definecolor{orange}{rgb}{0.8,0.4,0}
\definecolor{mylink}{RGB}{18,68,115}

\hypersetup{letterpaper,bookmarksopen,bookmarksnumbered,
pdfpagemode=UseOutlines,
colorlinks=true,
linkcolor=mylink,
anchorcolor=blue,
citecolor=mylink,
filecolor=blue,
menucolor=blue,
urlcolor=mylink,
}

\usepackage[utf8]{inputenc} 
\usepackage[T1]{fontenc}    
\usepackage{hyperref}       
\usepackage{url}            
\usepackage{booktabs}       
\usepackage{amsfonts}       
\usepackage{nicefrac}       
\usepackage{microtype}      

\usepackage{graphicx} 

\usepackage{amssymb}
\usepackage{amsmath}
\usepackage{amsthm}

\newcommand{\cirl}{{CIRL}}
\newcommand{\acirl}{{ACIRL}}

\newcommand{\expertdemon}{DBE}

\newcommand{\irl}{IRL}
\newcommand{\mdp}{{\sc mdp}}
\newcommand{\pomdp}{{\sc pomdp}}
\newcommand{\decpomdp}{{\sc d}ec-\pomdp}

\newcommand{\secref}[1]{Section~\ref{#1}}

\newcommand{\eqnref}[1]{Equation~\ref{#1}}
\newcommand{\figref}[1]{Figure~\ref{#1}}

\newcommand{\thmref}[1]{Theorem~\ref{#1}}

\newtheorem{thm}{Theorem}

\newtheorem{defn}{Definition}

\newtheorem{cor}{Corollary}

\newtheorem{rem}{Remark}

\newenvironment{tightlist}%
{\begin{list}{$\bullet$}{%
    \setlength{\topsep}{0in}
    \setlength{\partopsep}{0in}
    \setlength{\itemsep}{0in}
    \setlength{\parsep}{0in}
    \setlength{\leftmargin}{3.5em}
    \setlength{\rightmargin}{0in}
    \setlength{\itemindent}{-.1in}
}
}%
{\end{list}
}

\DeclareMathOperator*{\argmax}{argmax}

\newcommand{\principal}{\ensuremath{\mathbf{H}}}
\newcommand{\agent}{\ensuremath{\mathbf{R}}}
\newcommand{\coordinator}{\ensuremath{\mathbf{C}}}
\newcommand{\Rspace}{\ensuremath{\Theta}}
\newcommand{\Rparams}{\ensuremath{\theta}}
\newcommand{\statespace}{\ensuremath{\mathcal{S}}}
\newcommand{\state}{\ensuremath{s}}
\newcommand{\Pdots}{\ensuremath{P_0(\cdot, \cdot)}}
\newcommand{\Tdots}{\ensuremath{T(\cdot| \cdot, \cdot, \cdot)}}
\newcommand{\Rdots}{\ensuremath{R(\cdot, \cdot, \cdot; \cdot)}}

\newcommand{\feats}{\ensuremath{\phi}}
\newcommand{\nfeats}{\ensuremath{N_\feats}}

\newcommand{\Pactionspace}{\ensuremath{\mathcal{A}^\principal}}
\newcommand{\Paction}{\ensuremath{a^\principal}}
\newcommand{\Pdrule}{\ensuremath{\delta^\principal}}

\newcommand{\Pstrat}{\ensuremath{\pi^\principal}}
\newcommand{\Aaction}{\ensuremath{a^\agent}}
\newcommand{\Aactionspace}{\ensuremath{\mathcal{A}^\agent}}
\newcommand{\Astrat}{\ensuremath{\pi^\agent}}

\newcommand{\Cstrat}{\ensuremath{\pi^\coordinator}}

\newcommand{\agentBel}{{\ensuremath{b^\agent}}}

\newcommand{\expert}{{\ensuremath{\mathbf{E}}}}
\newcommand{\Estrat}{{\ensuremath{\pi^\mathbf{E}}}}

\newcommand{\bestresp}[1]{\ensuremath{\mathbf{br}(#1)}}

\title{Cooperative Inverse Reinforcement Learning}

\author{ {\bf Dylan Hadfield-Menell\thanks{\{dhm, anca, pabbeel, russell\}@cs.berkeley.edu} \hspace{30pt} Anca Dragan
 \hspace{30pt} Pieter Abbeel \hspace{30pt} Stuart Russell\vspace{5pt}} \\
Electrical Engineering and Computer Science \\
University of California at Berkeley\\
Berkeley, CA 94709 \\}

\begin{document}

\maketitle

\begin{abstract} 
For an autonomous system to be helpful to humans and to pose no
unwarranted risks, it needs to align its values with those of the
humans in its environment in such a way that its actions contribute to
the maximization of value for the humans. We propose a formal
definition of the value alignment problem as {\em cooperative inverse
reinforcement learning} (CIRL). A CIRL problem is a cooperative,
partial-information game with two agents, human and robot; both are
rewarded according to the human's reward function, but the robot does
not initially know what this is. In contrast to classical IRL, where
the human is assumed to act optimally in isolation, optimal CIRL
solutions produce behaviors such as active teaching, active learning,
and communicative actions that are more effective in achieving value
alignment. We show that computing optimal joint policies in CIRL games
can be reduced to solving a POMDP, prove that optimality in isolation
is suboptimal in CIRL, and derive an approximate CIRL algorithm.

\end{abstract} 

\section{Introduction}
\label{intro}
``{\em If we use, to achieve our purposes, a mechanical agency with
whose operation we cannot interfere effectively $\ldots$ we had better
be quite sure that the purpose put into the machine is the purpose
which we really desire.}'' So wrote Norbert
Wiener~(\citeyear{Wiener:1960}) in one of the earliest explanations of
the problems that arise when a powerful autonomous system operates
with an incorrect objective. This {\em value alignment} problem is far
from trivial. Humans are prone to mis-stating their objectives, which
can lead to unexpected implementations. In the myth of King Midas, the
main character learns that wishing for `everything he touches to turn
to gold' leads to disaster. In a reinforcement learning
context, \cite{Russell+Norvig:2010} describe a seemingly reasonable,
but incorrect, reward function for a vacuum robot: if we reward the
action of cleaning up dirt, the optimal policy causes the robot to
repeatedly dump and clean up the same dirt.

A solution to the value alignment problem has long-term implications
for the future of AI and its relationship to
humanity~\citep{bostrom2014superintelligence} and short-term utility
for the design of usable AI systems. Giving robots the right
objectives and enabling them to make the right trade-offs is crucial
for self-driving cars, personal assistants, and human--robot
interaction more broadly. 

The field of \emph{inverse reinforcement learning} or
IRL~\citep{Russell:1998,ng2000algorithms,abbeel2004apprenticeship} is
certainly relevant to the value alignment problem. An IRL algorithm
infers the reward function of an agent from observations of the
agent's behavior, which is assumed to be optimal (or approximately
so).  One might imagine that IRL provides a simple solution to the
value alignment problem: the robot observes human behavior, learns the
human reward function, and behaves according to that function. This
simple idea has two flaws. The first flaw is obvious: we don't want
the robot to adopt the human reward function as its own. For example,
human behavior (especially in the morning) often conveys a desire for
coffee, and the robot can learn this with IRL, but we don't want the robot to
want coffee! This flaw is easily fixed: we need to formulate the value
alignment problem so that the robot always has the fixed objective of
optimizing reward {\em for the human}, and becomes better able to do
so as it learns what the human reward function is.

The second flaw is less obvious, and less easy to fix. IRL assumes
that observed behavior is optimal in the sense that it accomplishes a
given task efficiently. This precludes a variety of useful teaching
behaviors. For example, efficiently making a cup of coffee, while the
robot is a passive observer, is a \emph{inefficient} way to teach a
robot to get coffee. Instead, the human should perhaps {\em explain}
the steps in coffee preparation and {\em show} the robot where the
backup coffee supplies are kept and what do if the coffee pot is left
on the heating plate too long, while the robot might {\em ask} what
the button with the puffy steam symbol is for and {\em try its hand}
at coffee making with guidance from the human, even if the first
results are undrinkable. None of these things fit in with the standard
IRL framework.




\noindent\textbf{Cooperative inverse reinforcement learning.} 
We propose, therefore, that value alignment should be formulated as a
{\em cooperative} and {\em interactive} reward maximization process.
More precisely, we define a \emph{cooperative inverse reinforcement
learning} (\cirl) game as a two-player game of partial information,
in which the ``human'', \principal, knows the reward function
(represented by a generalized parameter $\Rparams$), while the
``robot'', \agent, does not; the robot's payoff is exactly the human's
actual reward.  Optimal solutions to this game maximize human reward;
we show that solutions may involve active instruction by the human and
active learning by the robot.


\noindent \textbf{Reduction to POMDP and Sufficient Statistics.}
As one might expect, the structure of {\cirl} games is such that they
admit more efficient solution algorithms than are possible for general
partial-information games. Let $(\Pstrat, \Astrat)$ be a pair of
policies for human and robot, each depending, in general, on the
complete history of observations and actions. A policy pair yields an
expected sum of rewards for each player. \cirl{} games are
cooperative, so there is a well-defined optimal policy pair that
maximizes value.\footnote{A coordination problem of the type described
in \cite{boutilier1999sequential} arises if there are multiple optimal
policy pairs; we defer this issue to future work.}
In \secref{formulation} we reduce the problem of computing an optimal
policy pair to the solution of a (single-agent) \pomdp. This shows
that the robot's posterior over $\Rparams$ is a sufficient statistic,
in the sense that there are optimal policy pairs in which the robot's
behavior depends only on this statistic.  Moreover, the complexity of
solving the {\pomdp} is exponentially lower than the NEXP-hard bound
that \citep{bernstein2000complexity} obtained by reducing a \cirl{}
game to a general {\decpomdp}.

\noindent \textbf{Apprenticeship Learning and Suboptimality of IRL-Like Solutions.}
In \secref{acirl} we model apprenticeship
learning~\citep{abbeel2004apprenticeship} as a two-phase \cirl{}
game. In the first phase, the learning phase, both \principal{}
and \agent{} can take actions and this lets \agent{} learn
about \Rparams. In the second phase, the deployment phase, \agent{}
uses what it learned to maximize reward (without supervision
from \principal). We show that classic {\irl} falls out as the
best-response policy for \agent{} under the assumption that the
human's policy is ``demonstration by expert'' (DBE), i.e., acting
optimally {\em in isolation} as if no robot exists. But we show also
that this DBE/IRL policy pair is not, in general, optimal: even if the
robot expects expert behavior, demonstrating expert behavior is not
the best way to teach that algorithm. 

We give an algorithm that approximately computes \principal's best
response when \agent{} is running IRL under the assumption that
rewards are linear in \Rparams{} and state
features. \secref{gridworld} compares this best-response policy with
the DBE policy in an example game and provides empirical confirmation
that the best-response policy, which turns out to ``teach'' \agent{}
about the value landscape of the problem, is better than DBE. Thus,
designers of apprenticeship learning systems should \emph{expect} that
users will violate the assumption of expert demonstrations in order to
better communicate information about the objective.

\begin{figure}
\centering
\includegraphics[width=.32\textwidth]{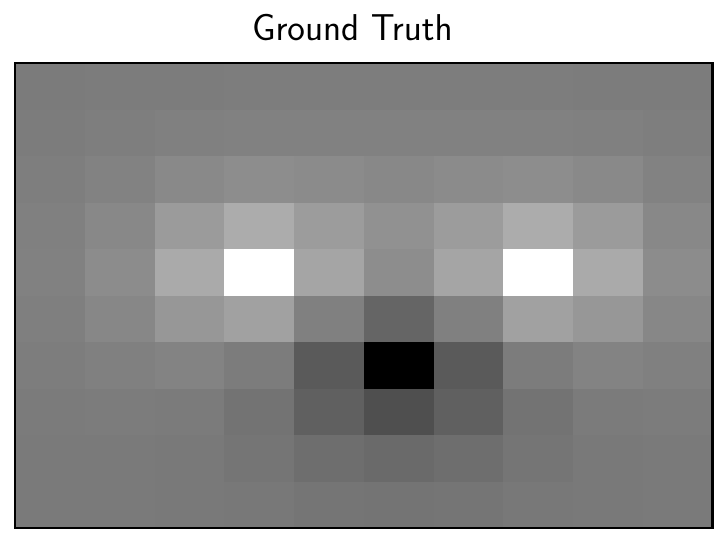}
\includegraphics[width=.32\textwidth]{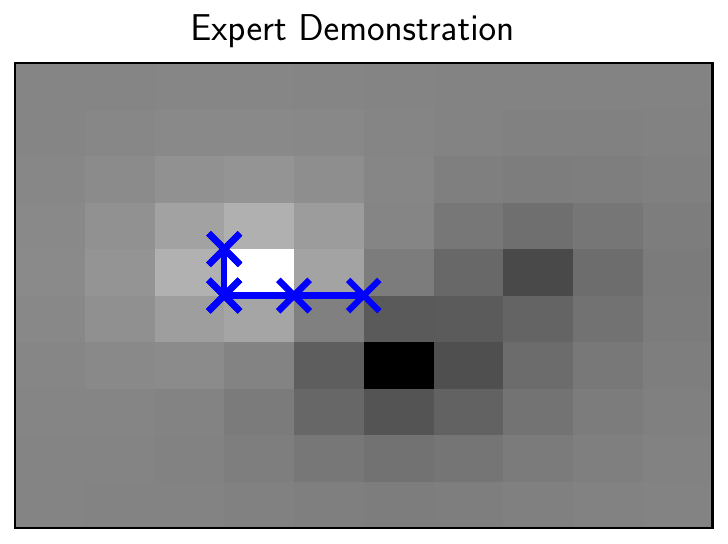}
\includegraphics[width=.32\textwidth]{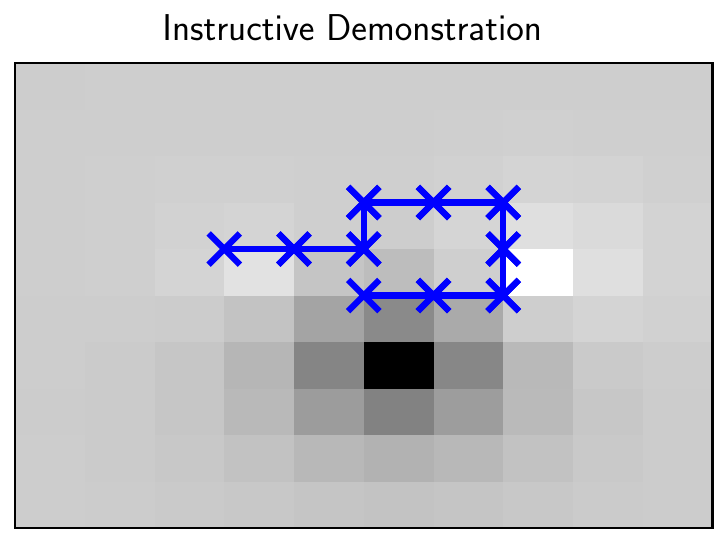}
\caption{The difference between demonstration-by-expert and
  instructive demonstration in the mobile robot navigation problem
  from \secref{gridworld}. Left: The ground truth reward
  function. Lighter grid cells indicates areas of higher
  reward. Middle: The demonstration trajectory generated by the expert
  policy, superimposed on the maximum a-posteriori reward function the
  robot infers. The robot successfully learns where the maximum reward
  is, but little else. Right: An instructive demonstration generated
  by the algorithm in \secref{br-match-feats} superimposed on the
  maximum a-posteriori reward function that the robot infers. This
  demonstration highlights both points of high reward and so the robot
  learns a better estimate of the reward.\vspace{-13pt}}
\label{fig-demo-example}
\end{figure}

\section{Related Work}
\label{related}

Our proposed model shares aspects with a variety of existing
models. We divide the related work into three categories: inverse
reinforcement learning, optimal teaching, and principal--agent
models.

\paragraph{Inverse Reinforcement Learning.}

\cite{ng2000algorithms} define \emph{inverse reinforcement learning}
(\irl) as follows: ``\textbf{Given} measurements of an
     [actor]'s behavior over time. $\ldots$ \textbf{Determine} the
     reward function being optimized.'' The key assumption
     \irl{} makes is that the observed behavior is optimal in the
     sense that the observed trajectory maximizes the sum of
     rewards. We call this the \emph{demonstration-by-expert}
     (\expertdemon) assumption. One of our contributions is to prove
     that this may be \emph{suboptimal} behavior in a CIRL game, as
     \principal{} may choose to accept less reward on a particular
     action in order to \emph{convey more information} to \agent. In
     CIRL the DBE assumption prescribes a fixed policy for
     \principal. As a result, many IRL algorithms can be derived as
     state estimation for a best response to different \Pstrat, where
     the state includes the unobserved reward parametrization
     \Rparams. 

\cite{ng2000algorithms}, \cite{abbeel2004apprenticeship}, and
\cite{Ratliff_ICML06} compute constraints that characterize the set of
reward functions so that the observed behavior maximizes reward. In
general, there will be many reward functions consistent with this
constraint. They use a max-margin heuristic to select a single reward
function from this set as their estimate. In \cirl, the constraints
they compute characterize \agent's belief about \Rparams{} under the
\expertdemon{} assumption. 

 \cite{ramachandran2007bayesian} and \cite{ziebart2008maximum}
 consider the case where \Pstrat{} is ``noisily expert,'' i.e.,
 \Pstrat is a Boltzmann distribution where actions or trajectories are
 selected in proportion to the exponent of their
 value. \cite{ramachandran2007bayesian} adopt a Bayesian approach and
 place an explicit prior on rewards. \cite{ziebart2008maximum} places
 a prior on reward functions indirectly by assuming a uniform prior
 over trajectories. In our model, these assumptions are variations of
 \expertdemon{} and both implement state estimation for a best
 response to the appropriate fixed \principal.

 \cite{natarajan2010multi} introduce an extension to \irl{} where
 \agent{} observes multiple actors that cooperate to maximize a common
 reward function. This is a different type of cooperation than we
 consider, as the reward function is common
 knowledge and \agent{} is a passive observer. \cite{waugh2011computational} and
 \cite{kuleshov2015inverse} consider the problem of inferring payoffs
 from observed behavior in a general (i.e., non-cooperative) game
 given observed behavior. It would be interesting to consider an
 analogous extension to \cirl, akin to mechanism design, in which
 \agent{} tries to maximize collective utility for a group of
 \principal s that may have competing objectives.

\cite{fern2014decision} consider a \emph{hidden-goal} \mdp, a special
case of a \pomdp{} where the goal is an unobserved part of the
state. This can be considered a special case of \cirl, where
\Rparams{} encodes a particular goal state. The frameworks share the
idea that \agent{} helps \principal. The key difference between the
models lies in the treatment of the human (the agent in their
terminology). \cite{fern2014decision} model the human as part of the
environment. In contrast, we treat \principal{} as an actor in a
decision problem that both actors collectively solve. This is crucial
to modeling the human's incentive to \emph{teach}.




\paragraph{Optimal Teaching.}

Because CIRL incentivizes the human to teach, as opposed to maximizing
reward in isolation, our work is related to optimal teaching: finding
examples that optimally train a learner~\citep{balbach2009recent,
goldman1993learning, goldman1995complexity}. The key difference is
that efficient learning is the \emph{objective} of optimal teaching,
while it emerges as a \emph{property} of optimal equilibrium behavior in \cirl.

\cite{cakmak2012algorithmic} consider an application of optimal
teaching where the goal is to teach the learner the reward function
for an \mdp. The teacher gets to pick initial states from which an
expert executes the reward-maximizing trajectory. The learner uses IRL
to infer the reward function, and the teacher picks initial states to
minimize the learner's uncertainty. In CIRL, this approach can be
characterized as an approximate algorithm for
\principal{} that greedily minimizes the entropy of \agent's belief.

Beyond teaching, several models focus on taking actions that convey
some underlying state, not necessarily a reward function. Examples
include finding a motion that best communicates an agent's
intention~\citep{dragan2013generating}, or finding a natural language
utterance that best communicates a particular
grounding~\citep{golland2010game}. All of these
approaches model the observer's inference process and compute actions
(motion or speech) that maximize the probability an observer infers
the correct hypothesis or goal. Our approximate solution to CIRL is
analogous to these approaches, in that we compute actions that are
informative of the correct reward function.

\paragraph{Principal--agent models.}
\label{pa-models}
Value alignment problems are not intrinsic to artificial agents.
\cite{kerr1975folly} describes a wide variety of misaligned incentives
in the aptly titled ``On the folly of rewarding A, while hoping for
B.'' In economics, this is known as the principal--agent problem: the
principal (e.g., the employer) specifies incentives so that an agent
(e.g., the employee) maximizes the principal's
profit~\citep{jensen1976theory}.

Principal--agent models study the problem of generating appropriate
incentives in a non-cooperative setting with asymmetric information.
In this setting, misalignment arises because the agents that economists
model are people and intrinsically have their own desires. In AI,
misalignment arises entirely from the information asymmetry between the
principal and the agent; if we could characterize the correct reward
function, we could program it into an artificial
agent. \cite{gibbons1998incentives} provides a useful survey of
principal--agent models and their
applications.

\section{Cooperative Inverse Reinforcement Learning}
\label{formulation}
This section formulates CIRL as a two-player Markov game with
identical payoffs, reduces the problem of computing an optimal
policy pair for a CIRL game to solving a POMDP, and characterizes \emph{apprenticeship learning} as a subclass of CIRL
games.

\subsection{CIRL Formulation}
\begin{defn}
A \emph{cooperative inverse reinforcement learning} (\cirl) game $M$
is a two-player Markov game with identical payoffs between a human or
principal, \principal, and a robot or agent, \agent. The game is
described by a tuple,
$M~=~\langle\statespace, \{\Pactionspace, \Aactionspace\}, \Tdots, \{\Rspace, \Rdots\}, \Pdots, \gamma \rangle,$
with the following definitions:
\begin{tightlist}
 \item[\statespace] a set of world states: $\state \in \statespace$.
 \item[\Pactionspace] a set of actions for \principal: $\Paction \in \Pactionspace$.
 \item[\Aactionspace] a set of actions for \agent: $\Aaction \in \Aactionspace$.
 \item[\Tdots] a conditional distribution on the next world state,
 given previous state and action for both agents: $T(\state'
 | \state, \Paction, \Aaction)$.

 \item[\Rspace] a set of possible static reward parameters, only observed by
   \principal: $\Rparams \in \Rspace$.

 \item[\Rdots] a parameterized reward function that maps world states,
   joint actions, and reward parameters to real numbers.
   $R: \statespace \times \Pactionspace \times \Aactionspace
   \times \Rspace \rightarrow \mathbb{R}.$

 \item[\Pdots] a distribution over the initial state, represented
 as tuples: $P_0(\state_0, \Rparams)$
 \item[$\gamma$] a discount factor: $\gamma \in [0, 1]$.
\end{tightlist}

\end{defn}

We write the reward for a state--parameter pair as $R(\state,
\Paction, \Aaction; \Rparams)$ to distinguish the static reward 
parameters $\theta$ from the changing world state $s$. The game
proceeds as follows. First, the initial state, a tuple
$(\state, \Rparams),$ is sampled from $P_0$. \principal{}
observes \Rparams, but \agent{} does not. This observation model
captures the notion that only the human knows the reward function,
while both actors know a prior distribution over possible reward
functions. At each timestep $t$, \principal{} and \agent{} observe the
current state $\state_t$ and select their actions
$\Paction_t, \Aaction_t.$ Both actors receive reward
$r_t~=~R(\state_t,~\Paction_t,~\Aaction_t;~\Rparams)$ and observe each
other's action selection.
A state for the next timestep is sampled
from the transition distribution, $\state_{t+1} \sim P_T(\state' |
\state_t, \Paction_t, \Aaction_t)$, and the process
repeats. 

Behavior in a \cirl{} game is defined by a pair of policies,
$(\Pstrat, \Astrat)$, that determine action selection for \principal{}
and \agent{} respectively. In general, these policies can be arbitrary
functions of their observation histories; $\Pstrat:
\left[{\Pactionspace} \times {\Aactionspace}
  \times \statespace\right]^* \times \Rspace
\rightarrow \Pactionspace, \Astrat: \left[{\Pactionspace} \times
  {\Aactionspace} \times \statespace\right]^*
\rightarrow \Aactionspace .$ The optimal joint policy is the policy
that maximizes \emph{value}. The value of a state is the expected sum
of discounted rewards under the initial distribution of reward
parameters and world states.

\begin{rem}
A key property of CIRL is that the human and the robot get rewards
determined by the same reward function. This incentivizes the human to
teach and the robot to learn without explicitly encoding these as
objectives of the actors.
\end{rem}

\subsection{Structural Results for Computing Optimal Policy Pairs}
\label{coord-pomdp}
The analogue in CIRL to computing an optimal policy for an MDP is the
problem of computing an optimal policy pair. This is a pair of
policies that maximizes the expected sum of discounted rewards. This
is not the same as `solving' a CIRL game, as a real world
implementation of a CIRL agent must account for coordination problems
and strategic uncertainty~\citep{boutilier1999sequential}. The optimal
policy pair represents the best \principal{} and \agent{} can do if
they can coordinate perfectly before \principal{}
observes \Rparams. Computing an optimal joint policy for a cooperative
game is the solution to a \emph{decentralized-partially observed
Markov decision process} (\decpomdp). Unfortunately, \decpomdp s are
NEXP-complete~\citep{bernstein2000complexity} so general \decpomdp{}
algorithms have a computational complexity that is doubly
exponential. Fortunately, \cirl{} games have special structure that
reduces this complexity.

\cite{nayyar2013decentralized} shows that a \decpomdp{} can be reduced
to a \emph{coordination}-\pomdp. The actor in this \pomdp{} is a
coordinator that observes all common observations and specifies a
policy for each actor. These policies map each actor's private
information to an action. The structure of a CIRL game implies that
the private information is limited to \principal's initial observation
of \Rparams. This allows the reduction to a coordination-\pomdp{} to
preserve the size of the (hidden) state space, making the problem
easier.


\begin{thm}
\label{thm-equiv}
Let $M$ be an arbitrary \cirl{} game with state space \statespace{}
and reward space \Rspace. There exists a (single-actor) \pomdp{}
$M_\coordinator$ with (hidden) state space $\mathcal{S}_\coordinator$
such that $|\mathcal{S}_\coordinator| = |\mathcal{S}| \cdot |\Rspace|$ and,
for any policy pair in $M$, there is a policy in $M_\coordinator$
that achieves the same sum of discounted rewards.
\end{thm}

Theorem proofs can be found in the supplementary material. An
immediate consequence of this result is that \agent's belief about
\Rparams{} is a sufficient statistic for optimal behavior. 

\begin{cor}
\label{cor-sufficient-stats}
Let $M$ be a \cirl{} game. There exists an optimal policy pair $({\Pstrat}^*,
{\Astrat}^*)$ that only depends on the current state and \agent's
belief. 
\end{cor} 

\begin{rem}
In a general \decpomdp, the hidden state for the coordinator-\pomdp{}
includes each actor's history of observations. In \cirl, \Rparams{} is
the only private information so we get an exponential decrease in the
complexity of the reduced problem. This allows one to apply general
\pomdp{} algorithms to compute optimal joint policies in \cirl.
\end{rem}

It is important to note that the reduced problem may still be very
challenging. \pomdp s are difficult in their own right and the reduced
problem still has a much larger action space. That being said, this
reduction is still useful in that it characterizes optimal joint
policy computation for \cirl{} as significantly easier than \decpomdp
s. Furthermore, this theorem can be used to justify approximate
methods (e.g., iterated best response) that only depend on \agent's
belief state.


\subsection{Apprenticeship Learning as a Subclass of CIRL Games}
\label{acirl}

A common paradigm for robot learning from humans
is \emph{apprenticeship learning}. In this paradigm, a human gives
demonstrations to a robot of a sample task and the robot is asked to
imitate it in a subsequent task. In what follows, we formulate
apprenticeship learning as turn-based CIRL with a learning phase and a
deployment phase. We characterize IRL as the best response (i.e., the
policy that maximizes reward given a fixed policy for the other
player) to a demonstration-by-expert policy for \principal. We also
show that this policy is, in general, \emph{not part of an optimal
joint policy} and so IRL is generally a suboptimal approach to
apprenticeship learning.

\begin{defn} (ACIRL) An \emph{apprenticeship cooperative inverse reinforcement learning}
(\acirl) game is a turn-based \cirl{} game with two phases: a learning
phase where the human and the robot take turns acting, and a
deployment phase, where the robot acts independently.
\end{defn}

 \paragraph{Example.} Consider an example apprenticeship task
where \agent{} needs to help \principal{} make office
supplies. \principal{} and \agent{} can make paperclips and staples
and the unobserved \Rparams{} describe \principal's preference for
paperclips vs staples. We model the problem as an \acirl{} game in which
the learning and deployment phase each consist of an individual
action. The world state in this problem is a tuple $(p_s, q_s, t)$
where $p_s$ and $q_s$ respectively represent the number of paperclips
and staples \principal{} owns. $t$ is the round number. An action is a
tuple $(p_a, q_a)$ that produces $p_a$ paperclips and $q_a$
staples. The human can make 2 items total:
$\Pactionspace~=~\{(0,2),(1,1),(2,0)\}.$ The robot has different
capabilities. It can make 50 units of each item or it can choose to
make 90 of a single item: $\Aactionspace~=~\{(0,90),(50,50),(90,0)\}.$
We let $\Rspace = [0, 1]$ and define $R$ so that \Rparams{} indicates
the relative preference between paperclips and staples:$R(s, (p_a,
q_a); \Rparams) = \Rparams p_a + (1-\Rparams) q_a.$ \agent's action is
ignored when $t=0$ and \principal's is ignored when $t=1$. At $t=2$,
the game is over, so the game transitions to a sink state, $(0, 0,
2)$. 

 \paragraph{Deployment phase --- maximize mean reward estimate.} It is
 simplest to analyze the deployment phase first. \agent{} is the only
 actor in this phase so it get no more observations of its reward. We
 have shown that \agent's belief about \Rparams{} is a sufficient
 statistic for the optimal policy. This belief about \Rparams{}
 induces a distribution over MDPs. A straightforward extension of a
 result due to \cite{ramachandran2007bayesian} shows that \agent's
 optimal deployment policy maximizes reward for the mean reward
 function.

\begin{thm}
\label{thm-max-reward}
Let $M$ be an \acirl{} game. In the deployment phase, the optimal
policy for \agent{} maximizes reward in the \mdp{} induced by the
mean \Rparams{} from  \agent's belief.
\end{thm} 

In our example, suppose that \Pstrat{} selects $(0, 2)$ if
$\Rparams \in [0, \frac{1}{3})$, $(1, 1)$ if $\Rparams \in
[\frac{1}{3}, \frac{2}{3}]$ and $(2, 0)$ otherwise. \agent{} begins
with a uniform prior on \Rparams{} so observing, e.g., $\Paction = (0,
2)$ leads to a posterior distribution that is uniform on $[0,
\frac{1}{3})$. \thmref{thm-max-reward} shows that the optimal action maximizes
reward for the mean \Rparams{} so an optimal \agent{} behaves as
though $\Rparams{} = \frac{1}{6}$ during the deployment phase.

 \paragraph{Learning phase --- expert demonstrations are not optimal.}
A wide variety of apprenticeship learning approaches assume that
demonstrations are given by an expert. We say that \principal{}
satisfies the \emph{demonstration-by-expert} (\expertdemon) assumption
in \acirl{} if she greedily maximizes immediate reward on her
turn. This is an `expert' demonstration because it demonstrates a
reward maximizing action but does not account for that action's impact
on \agent's belief. We let \Estrat{} represent the DBE policy. 

\thmref{thm-max-reward} enables us to characterize the best response for \agent{} when $\Pstrat=\Estrat$: use \irl{} to compute the
posterior over \Rparams{} during the learning phase and then act to
maximize reward under the mean \Rparams{} in the deployment phase. We
can also analyze the \expertdemon{} assumption itself. In particular,
we show that \Estrat{} is not \principal's best response
when \Astrat{} is a best response to \Estrat.

\begin{thm}
\label{irl-sub-opt}
There exist \acirl{} games where the best-response for \principal{}
to \Astrat{} violates the expert demonstrator assumption. In other
words, if $\bestresp{\pi}$ is the best response to $\pi$, then
$\bestresp{\bestresp{\Estrat}} \neq \Estrat$.
\end{thm} 


The supplementary material proves this theorem by computing the
optimal equilibrium for our example. In that equilibrium, \principal{}
selects $(1, 1)$ if $\theta \in [\frac{41}{92}, \frac{51}{92}]$. In
contrast, \Estrat{} only chooses $(1, 1)$ if $\theta = 0.5$. The
change arises because there are situations (e.g., $\theta = 0.49$)
where the immediate loss of reward to \principal{} is worth the
improvement in \agent's estimate of \Rparams.

\begin{rem}
We should expect experienced users of apprenticeship learning systems
to present demonstrations optimized for fast learning rather than
demonstrations that maximize reward.
\end{rem}

Crucially, the demonstrator is incentivized to deviate from \agent's
assumptions.  This has implications for the design and analysis of
apprenticeship systems in robotics. Inaccurate assumptions about user
behavior are notorious for exposing bugs in software systems (see,
e.g., \cite{leveson1993investigation}).

\subsection{Generating Instructive Demonstrations}
\label{br-match-feats}
Now, we consider the problem of computing \principal's best response
when \agent{} uses IRL as a state estimator. For our toy example, we
computed solutions exhaustively, for realistic problems we need a
more efficient approach. \secref{coord-pomdp} shows that this can be
reduced to an POMDP where the state is a tuple of world state, reward
parameters, and \agent's belief. While this is easier than solving a
general \decpomdp, it is a computational challenge. If we restrict our
attention to the case of linear reward functions we can develop an
efficient algorithm to compute an approximate best response.

Specifically, we consider the case where the reward for a state
$(s, \Rparams)$ is defined as a linear combination of state features
for some feature function $\feats: R(s, \Paction, \Aaction; \Rparams)
= \feats(s)^\top\Rparams$. Standard results from the \irl{} literature
show that policies with the same expected feature counts have the same
value~\citep{abbeel2004apprenticeship}. Combined
with \thmref{thm-max-reward}, this implies that the optimal \Astrat{} under
the \expertdemon{} assumption computes a policy that matches the
observed feature counts from the learning phase.

This suggests a simple approximation scheme. To compute a
demonstration trajectory $\tau^\principal$, first compute the feature
counts \agent{} would observe in expectation from the true \Rparams{}
and then select actions that maximize similarity to these target
features. If $\feats_\Rparams$ are the expected feature counts induced
by \Rparams{} then this scheme amounts to the following decision rule:\vspace{-10pt}

\begin{equation}
\tau^\principal \leftarrow \underset{\tau}{\argmax}\  \feats(\tau)^\top \Rparams - \eta ||\feats_{\Rparams} - \feats(\tau)||^2.
\end{equation}

\vspace{-5pt} This rule selects a trajectory that trades off between the sum of
rewards $\feats(\tau)^\top\Rparams$ and the feature dissimilarity
$||\feats_{\Rparams} - \feats(\tau)||^2$. Note that this is generally
distinct from the action selected by the demonstration-by-expert
policy. The goal is to match the expected sum of features under
a \emph{distribution} of trajectories with the sum of features from
a \emph{single} trajectory.  The correct measure of feature similarity
is \emph{regret}: the difference between the reward \agent{} would
collect if it knew the true \Rparams{} and the reward \agent{}
actually collects using the inferred \Rparams. Computing this
similarity is expensive, so we use an $\ell_2$ norm as a proxy measure
of similarity.

\section{Experiments}
\label{gridworld}

\begin{figure}
\centering
\includegraphics[width=\textwidth]{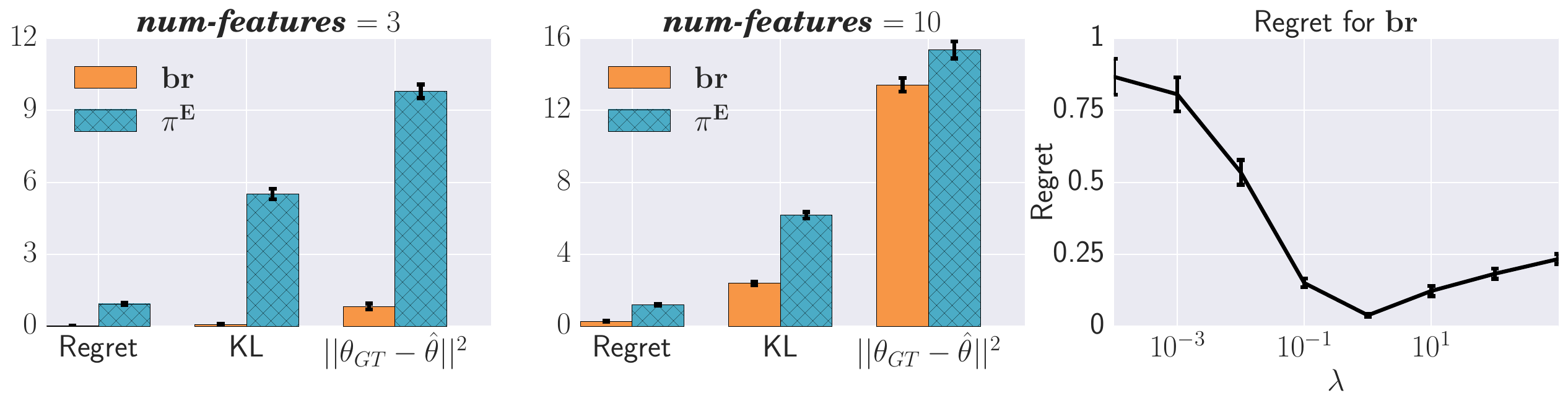}
\caption{{\small Left, Middle: Comparison of `expert' demonstration (\Estrat) with
  `instructive' demonstration ($\mathbf{br}$). Lower numbers are
  better. Using the best response causes \agent{} to infer a better
  distribution over \Rparams{} so it does a better job of maximizing
  reward. Right: The regret of the instructive demonstration policy as
  a function of how optimal \agent{} expects \principal{} to
  be. $\lambda=0$ corresponds to a robot that expects purely random
  behavior and $\lambda=\infty$ corresponds to a robot that expects
  optimal behavior. Regret is minimized for an intermediate value of
  $\lambda$: if $\lambda$ is too small, then \agent{} learns nothing
  from its observations; if $\lambda$ is too large, then \agent{}
  expects many values of \Rparams{} to lead to the same trajectory so
  \principal{} has no way to differentiate those reward functions. \vspace{-10pt}}}
\label{fig-results}
\end{figure}

\subsection{Cooperative Learning for Mobile Robot Navigation}

Our experimental domain is a 2D navigation problem on a discrete
grid. In the learning phase of the game, \principal{} teleoperates a
trajectory while \agent{} observes. In the deployment phase, \agent{}
is placed in a random state and given control of the robot. We use a
finite horizon $H$, and let the first $\frac{H}{2}$ timesteps be the
learning phase. There are \nfeats{} state features defined as radial
basis functions where the centers are common knowledge. Rewards are
linear in these features and \Rparams. The initial world state is in
the middle of the map. We use a uniform distribution on $[-1,
1]^{\nfeats}$ for the prior on \Rparams. Actions move in one of the
four cardinal directions $\{N, S, E, W\}$ and there is an additional
no-op $\emptyset$ that each actor executes deterministically on the
other agent's turn.



\figref{fig-demo-example} shows an example comparison between
demonstration-by-expert and the approximate best response policy in
\secref{br-match-feats}. The leftmost image is the ground truth reward
function. Next to it are demonstration trajectories produce by these
two policies. Each path is superimposed on the maximum a-posteriori
reward function the robot infers from the demonstration. We can see
that the demonstration-by-expert policy immediately goes to the
highest reward and stays there. In contrast, the best response policy
moves to both areas of high reward. The robot reward function the
robot infers from the best response demonstration is much more
representative of the true reward function, when compared with the
reward function it infers from demonstration-by-expert.

\subsection{Demonstration-by-Expert vs Best Responder}
\label{exp-1}


\noindent \textbf{Hypothesis. }When \agent{} plays an \irl{} algorithm
that matches features, \principal{} prefers the best response policy
from \secref{br-match-feats} to \Estrat{}: the best response policy
will significantly outperform the DBE policy.

 \textbf{Manipulated Variables.} Our experiment consists of 2
 factors: \textbf{\emph{\principal-policy}}
 and \textbf{\emph{num-features}}. We make the assumption
 that \agent{} uses an \irl{} algorithm to compute its estimate
 of \Rparams{} during learning and maximizes reward under this
 estimate during deployment. We use
 Maximum-Entropy \irl~\citep{ziebart2008maximum} to implement \agent's
 policy. \textbf{\emph{\principal-policy}} varies \principal's
 strategy \Pstrat and has two levels: demonstration-by-expert
 (\Estrat) and best-responder ($\mathbf{br}$). In the \Estrat{}
 level \principal{} maximizes reward during the demonstration. In the
 $\mathbf{br}$ level \principal{} uses the approximate algorithm
 from \secref{br-match-feats} to compute an approximate best
 response to \Astrat. The trade-off between reward and communication
 $\eta$ is set by cross-validation before the game
 begins. The \textbf{\emph{num-features}} factor varies the
 dimensionality of \feats across two levels: 3 features and 10
 features. We do this to test whether and how the difference between
 experts and best-responders is affected by dimensionality. We use a
 factorial design that leads to $4$ distinct conditions. We test each
 condition against a random sample of $N=500$ different reward
 parameters. We use a within-subjects design with respect to the
 the \textbf{\emph{\principal}-policy} factor so the same reward
 parameters are tested for \Estrat{} and $\mathbf{br}$. 

 \textbf{Dependent Measures.} We use the regret with respect to a
 fully-observed setting where the robot knows the ground
 truth \Rparams{} as a measure of performance. We let $\hat{\Rparams}$
 be the robot's estimate of the reward parameters and let
 $\Rparams_{GT}$ be the ground truth reward parameters. The primary
 measure is the \emph{regret} of \agent's policy: the difference
 between the value of the policy that maximizes the inferred reward
 $\hat{\Rparams}$ and the value of the policy that maximizes the true
 reward $\Rparams_{GT}$. We also use two secondary measures. The first
 is the KL-divergence between the maximum-entropy trajectory
 distribution induced by $\hat{\Rparams}$ and the maximum-entropy
 trajectory distribution induced by $\Rparams$. Finally, we use the
 $\ell_2$-norm between the vector or rewards defined by
 $\hat{\Rparams}$ and the vector induced by $\Rparams_{GT}$.


 \textbf{Results.}  There was relatively little correlation between
 the measures (Cronbach's $\alpha$ of .47), so we ran a factorial
 repeated measures ANOVA for each measure. Across all measures, we
 found a significant effect for \textbf{\emph{\principal-policy}},
 with $\mathbf{br}$ outperforming \Estrat{} on all measures as we
 hypothesized (all with $F > 962$, $p < .0001$). We did find an
 interaction effect with \textbf{\emph{num-features}} for
 KL-divergence and the $\ell_2$-norm of the reward vector but post-hoc
 Tukey HSD showed $\mathbf{br}$ to always outperform \Estrat. The
 interaction effect arises because the gap between the two levels
 of \textbf{\emph{\principal-policy}} is larger with fewer reward
 parameters; we interpret this as evidence that
 $\textbf{\emph{num-features}}=3$ is an easier teaching problem
 for \principal. \figref{fig-results} (Left, Middle) shows the
 dependent measures from our experiment.

\subsection{Varying \agent's Expectations}
Maximum-Entropy IRL includes a free parameter $\lambda$ that controls
how optimal \agent{} expects \principal{} to behave. If $\lambda = 0$,
\agent{} will update its belief as if \principal's observed behavior is
 \emph{independent} of her preferences \Rparams. If
 $\lambda=\infty$, \agent{} will update its belief as if \principal's
 behavior is \emph{exactly} optimal. We ran a followup experiment to
 determine how varying $\lambda$ changes the regret of the
 $\mathbf{br}$ policy. 

Changing $\lambda$ changes the forward model in \agent's belief
 update: the mapping \agent{} hypothesizes between a given reward
 parameter \Rparams{} and the observed feature counts
 $\phi_\theta$. This mapping is many-to-one for extreme values of
 $\lambda$. $\lambda \approx 0$ means that all values of \Rparams{}
 lead to the same expected feature counts because trajectories are
 chosen uniformly at random. Alternatively, $\lambda >> 0$ means that
 almost all probability mass falls on the optimal trajectory and many
 values of \Rparams{} will lead to the same optimal trajectory. This
 suggests that it is easier for \principal{} to differentiate
 different values of \Rparams{} if \agent{} assumes she is noisily
 optimal, but only up until a maximum noise
 level. \figref{fig-results} plots regret as a function of $\lambda$
 and supports this analysis: \principal{} has less regret for
 intermediate values of $\lambda$.









\section{Conclusion and Future Work}
\label{future}
In this work, we presented a game-theoretic model for cooperative
learning, \cirl. Key to this model is that the robot \emph{knows} that
it is in a shared environment and is attempting to maximize the
human's reward (as opposed to estimating the human's reward function
and adopting it as its own). This leads to cooperative learning
behavior and provides a framework in which to design HRI algorithms
and analyze the incentives of both actors in a reward learning
environment.

We reduced the problem of computing an optimal policy pair to solving
a \pomdp. This is a useful theoretical tool and can be used to design
new algorithms, but it is clear that optimal policy pairs are only
part of the story. In particular, when it performs a centralized
computation, the reduction assumes that we can effectively program
both actors to follow a set coordination policy. This is clearly
infeasible in reality, although it may nonetheless be helpful in
training humans to be better teachers. An important avenue for future
research will be to consider the coordination problem: the process by
which two independent actors arrive at policies that are mutual best
responses. Returning to Wiener's warning, we believe that the best
solution is not to put a specific purpose into the machine at all, but
instead to design machines that provably converge to the right purpose
as they go along.

\section*{Acknowledgments}
This work was supported by the DARPA Simplifying Complexity in
Scientific Discovery (SIMPLEX) program, the Berkeley Deep Drive
Center, the Center for Human Compatible AI, the Future of Life
Institute, and the Defense Sciences Office contract
N66001-15-2-4048. Dylan Hadfield-Menell is also supported by a NSF
Graduate Research Fellowship.

\newpage
\section{Appendix: Supplementary Material \& Proofs}

This appendix contains the supplementary material and proofs from the NeurIPS version of the paper. It is somewhat redundant with the original text to make it more self-contained.

\setcounter{defn}{0}
\setcounter{rem}{0}
\setcounter{thm}{0}
\setcounter{cor}{0}

\subsection{CIRL Formulation}
This section formulates CIRL as a two-player Markov game with
identical payoffs, reduces the problem of computing an optimal
equilibrium for a CIRL game to solving a POMDP, and
characterizes \emph{apprenticeship learning} as a subclass of CIRL
games.

\begin{defn}
A \emph{cooperative inverse reinforcement learning} (\cirl) game $M$
is a two-player Markov game with identical payoffs between a human or
principal, \principal, and a robot or agent, \agent. The game is
described by a tuple,
$M~=~\langle\statespace, \{\Pactionspace, \Aactionspace\}, \Tdots, \{\Rspace, \Rdots\}, \Pdots, \gamma \rangle,$
with the following definitions:
\begin{tightlist}
 \item[\statespace] a set of world states: $\state \in \statespace$.
 \item[\Pactionspace] a set of actions for \principal: $\Paction \in \Pactionspace$.
 \item[\Aactionspace] a set of actions for \agent: $\Aaction \in \Aactionspace$.
 \item[\Tdots] a conditional distribution on the next world state,
 given previous state and action for both agents: $T(\state'
 | \state, \Paction, \Aaction)$.

 \item[\Rspace] a set of possible static reward parameters, only observed by
   \principal: $\Rparams \in \Rspace$.

 \item[\Rdots] a parameterized reward function that maps world states,
   joint actions, and reward parameters to real numbers.
   $R: \statespace \times \Pactionspace \times \Aactionspace
   \times \Rspace \rightarrow \mathbb{R}.$

 \item[\Pdots] a distribution over the initial state, represented
 as tuples: $P_0(\state_0, \Rparams)$
 \item[$\gamma$] a discount factor: $\gamma \in [0, 1]$.
\end{tightlist}

\end{defn}

We write the reward for a state--parameter pair as $R(\state,
\Paction, \Aaction; \Rparams)$ to distinguish the static reward 
parameters $\theta$ from the changing world state $s$. 

The game proceeds as follows. First, the initial state, a tuple
$(\state, \Rparams),$ is sampled from $P_0$. \principal{} observes
\Rparams. This parameter represents the human's internal reward function. This observation models that only the human knows the reward function, while both actors know a prior distribution over possible reward functions. At
each timestep $t$, \principal{} and \agent{} observe the current state
$\state_t$ and select their actions $\Paction_t, \Aaction_t.$ Both
actors receive reward
$r_t~=~R(\state_t,~\Paction_t,~\Aaction_t;~\Rparams)$ and observe each
other's action selection.
A state for the next timestep is sampled
from the transition distribution, $\state_{t+1} \sim P_T(\state' |
\state_t, \Paction_t, \Aaction_t)$, and the process
repeats. 

Behavior in a \cirl{} game is defined by a pair of policies,
$(\Pstrat, \Astrat)$, that determine action selection for \principal{}
and \agent{} respectively. In general, these policies can be arbitrary
functions of their observation histories; $\Pstrat:
\left[{\Pactionspace} \times {\Aactionspace}
  \times \statespace\right]^* \times \Rspace
\rightarrow \Pactionspace, \Astrat: \left[{\Pactionspace} \times
  {\Aactionspace} \times \statespace\right]^*
\rightarrow \Aactionspace .$ The optimal joint policy is the policy
that maximizes \emph{value}. The value of a state is the expected sum
of discounted rewards under the initial distribution of reward
parameters and world states.

\begin{rem}
A key property of CIRL is that the human and the robot get rewards
determined by the same reward function. This incentivizes the human to
teach and the robot to learn without explicitly encoding these as
objectives of the actors.
\end{rem}

\subsection{Structural Results for Optimal Equilibrium Computation}
The analogue in CIRL to computing an optimal policy for an MDP is the
problem of computing an optimal policy pair. This is a pair of
policies that maximizes the expected sum of discounted rewards. This
is not the same as `solving' a CIRL game, as a real world
implementation of a CIRL agent must account for coordination problems
and strategic uncertainty~\citep{boutilier1999sequential}. The optimal
policy pair represents the best \principal{} and \agent{} can do if
they can coordinate perfectly before \principal{} observes \Rparams.

Computing an optimal joint policy for a cooperative game is the
solution to a \emph{decentralized-partially observed Markov decision
  process} (\decpomdp). Unfortunately, \decpomdp s are
NEXP-complete~\citep{bernstein2000complexity} so general \decpomdp{}
algorithms have a computational complexity that is doubly
exponential. Fortunately, \cirl{} games have special structure that
makes optimal equilibrium computation more efficient.

\cite{nayyar2013decentralized} shows that a \decpomdp{} can be reduced
to a \emph{coordination}-\pomdp. The actor in this \pomdp{} is a
coordinator that observes all common observations and specifies a
policy for each actor. These policies map each actor's private
information to an action. The structure of a CIRL game implies that
the private information is limited to \principal's initial observation
of \Rparams. This allows the reduction to a coordination-\pomdp{} to
preserve the size of the (hidden) state space, making the problem
easier.

\begin{defn} Let $M$ be a \cirl{} game between
   \principal{} and \agent. The corresponding \emph{coordination
     \pomdp} $M_\coordinator$ is a \pomdp{} where the single actor is
   a coordinator \coordinator. States are tuples of world state and
   reward parameters: $\mathcal{S}_c = \statespace
   \times \Rspace$. The initial state distribution places the same
   distribution on $\statespace \times \Rspace$ as
   $P_0$. \coordinator's actions are tuples (\Pdrule, \Aaction) that
   specify an action for \agent{} and a decision rule for \principal{}
   that maps its private information (\Rparams) to an action
   $\Pdrule: \Rspace \rightarrow \Pactionspace.$ \coordinator{}
   observes \principal's action and the world state. Transitions are
   defined analogously to those in $M$.
\end{defn}

\begin{thm}
Let $M$ be an arbitrary \cirl{} game with state space \statespace{}
and reward space \Rspace. There exists a (single-actor) \pomdp{}
$M_\coordinator$ with (hidden) state space $\mathcal{S}_\coordinator$
such that $|\mathcal{S}_\coordinator| = |\mathcal{S}| \cdot |\Rspace|$ and,
for any policy pair in $M$, there is a policy in $M_\coordinator$
that achieves the same sum of discounted rewards.
\end{thm}

\begin{proof}
We take $M_\coordinator$ to be the coordination POMDP associated
associated with $M$. The second component of \coordinator's action is
an action for \agent. \agent{} has no private observations, so for any
policy \Astrat{} \agent{} could choose to follow, \coordinator can
match it by simulating \Astrat{} and outputting the corresponding
action. Similarly, \coordinator{} only observes common observations,
so \agent{} can implement any coordinator strategy by simulating
\coordinator{} and directly executing the appropriate action.

By a similar arguement, \principal{} can also simulate any given
\Cstrat{} to compute her decision rule \Pdrule, and then execute the
corresponding action. To see that there is a \Cstrat{} that can
reproduce the behavior of any \Pstrat, let $h$ be the
action-observation history for \principal. \coordinator{} can choose
the following decision rule
$$\Pdrule(\Rparams) = \Pstrat(\Rparams; h)$$ to produce the same
behavior.
\end{proof}

\begin{cor}
Let $M$ be a \cirl{} game. There exist optimal policies $({\Pstrat}^*,
{\Astrat}^*)$ that only depend on the current state and \agent's
belief. 
\begin{align*}
&{\Pstrat}^*: \statespace \times \Delta_{\Rspace} \times \Rspace \rightarrow \Pactionspace, & {\Astrat}^*: \statespace \times \Delta_{\Rspace} \rightarrow \Aactionspace .\end{align*}
\end{cor}
\begin{proof}
\cite{smallwood1973optimal} showed that an optimal policy in a
\pomdp{} only depends on the belief state. \agent's belief uniquely
determines the belief for \coordinator. From this, an appeal to
\thmref{thm-equiv} shows the result.
\end{proof}

\subsection{Apprenticeship CIRL}

\paragraph{Example.} Consider an example apprenticeship task where \agent{} needs to help \principal{} make office supplies. \principal{} and \agent{} can make paperclips and staples and
the unobserved \Rparams{} describe \principal's preference for
paperclips vs staples. We model the problem as an \acirl{} in which
the learning and deployment phase each consist of an individual
action.

The world state in this problem is a tuple $(p_s, q_s, t)$ where $p_s$
and $q_s$ respectively represent the number of paperclips and
staples \principal{} owns. $t$ is the round number. An action is a
tuple $(p_a, q_a)$ that produces $p_a$ paperclips and $q_a$
staples. The human can make 2 items in total:
$\Pactionspace~=~\{(0,2),(1,1),(2,0)\}.$ The robot has different
capabilities. It can make 50 units of each item or it can choose to
make 90 of a single item: $\Aactionspace~=~\{(0,90),(50,50),(90,0)\}.$

We let $\Rspace = [0, 1]$ and define $R$ so that \Rparams{} indicates
the relative preference between paperclips and staples:$R(s, (p_a,
q_q); \Rparams) = \Rparams p_a + (1-\Rparams) q_a.$ \agent's action is
ignored when $t=0$ and \principal's is ignored when $t=1$. At $t=2$,
the game is over, so we transition to a sink state, $(0, 0,
2)$. Initially, there are no paperclips or staples, and we use a
uniform prior on \Rparams.

 \principal{} only acts in the initial state, so \Pstrat can be
entirely describe by a single decision rule $\Pdrule: [0,
1] \rightarrow \Pactionspace$.  \agent{} only observes one action
from \principal{} and so the reachable beliefs are in one-to-one
correspondence with \principal's actions. This lets us
characterize \agent's policy as
$\Astrat: \Pactionspace \rightarrow \Aactionspace.$

\begin{thm}
Let $M$ be an \acirl{} game. In the deployment phase, the optimal
policy for \agent{} maximizes reward in the \mdp{} induced by the
mean \Rparams.
\end{thm}
\begin{proof}
If \agent{} never observes another action from \principal{}, then
there are no common observations, so the coordination \pomdp{} has no
observations. The unobserved component of the state is static, so this
distribution does not change over time. This reduces the problem to
solving an \mdp{} under a fixed distribution over reward functions so
Theorem 3 from \cite{ramachandran2007bayesian} shows the result.
\end{proof}

The \expertdemon{} assumption in our example assumes that \principal{}
maximize reward in the first round. Let $\Rparams =
0.49$. \principal{} maximizes reward and chooses to make 0 paperclips
and 2 staples. \agent{} observes this and updates its belief (using
$\delta^\expert$ to define the observation distribution). In this
case, we get $\agentBel=\mathbf{Unif}([0, 0.5))$.  Given this
belief, \agent{} maximizes expected reward and chooses to make 0
paperclips and 90 staples. Thus, the expert decision rule $\delta^\expert$
and its \emph{best response} $\bestresp{\delta^\expert}$ are defined by

\begin{align}
\label{drule-E}
\delta^\expert (\Rparams) &= \left\{
\begin{array}{lr}
(0, 2) & \hspace{27pt}\Rparams < 0.5 \\ (1, 1) & \Rparams = 0.5 \\ (2, 0)
& \Rparams > 0.5
\end{array}
\right., \\
\bestresp{\delta^\expert}(\Paction) &= \left\{
\begin{array}{lr}
(0, 90) & \Paction = (0, 2) \\ (50, 50) & \Paction = (1, 1)\\ (90, 0) & \Paction = (2, 0)
\end{array}
\right. .
\end{align}



Note that when $\Rparams=0.49$ \principal{} would prefer \agent{} to
choose (50, 50). \principal{} is willing to forgo immediate reward
during the demonstration to communicate this to \agent: the best
response chooses $(1, 1)$ when $\Rparams=0.49$. This
leads to the following result.

\begin{thm}
There exist \acirl{} games where the best-response for \principal{}
to \Astrat{} violates the expert demonstrator assumption. In other
words, if $\bestresp{\pi}$ is the best response to $\pi$, then
$\bestresp{\bestresp{\Estrat}} \neq \Estrat$.
\end{thm}

\begin{proof}
Our office supply example gives a counter-example that shows the
 theorem. When \principal{} accounts for \agent's actions under
 $\bestresp{\delta^\expert}$, \principal{} is faced with a choice
 between 0 paperclips and 92 staples, 51 of each, or
 92 paperclips and 0 staples. It is straightforward to show that the
 optimal decision rule is given by

\begin{equation*}
\Pdrule(\Rparams) = \left\{
\begin{array}{lr}
(0, 2) & \Rparams < \frac{41}{92}\vspace{2pt} \\ (1, 1)
&\hspace{6pt} \frac{41}{92} \leq \Rparams \leq \frac{51}{92} \vspace{2pt}\\
(2, 0) & \Rparams > \frac{51}{92}
\end{array} \right. .
\end{equation*}

This is distinct from \eqnref{drule-E}, so we conclude the result.
\end{proof}

\newpage
\bibliography{biblio}
\bibliographystyle{icml2016}

\end{document}